\documentclass[10pt,twocolumn,letterpaper]{article}

\usepackage{cvpr}
\usepackage{times}
\usepackage{epsfig}
\usepackage{graphicx}
\usepackage{amsmath}
\usepackage{amssymb}
\usepackage[pagebackref=true,breaklinks=true,letterpaper=true,colorlinks,bookmarks=false]{hyperref}
\usepackage{amsfonts,amssymb}
\usepackage{verbatim}

\graphicspath{{figures/}}
\DeclareGraphicsExtensions{.png,.jpg,.eps,.pdf}

\usepackage[breaklinks=true,bookmarks=false]{hyperref}

\cvprfinalcopy 

\def\cvprPaperID{****} 

\begin{document}

\title{One-Shot Imitation Filming of Human Motion Videos}

\author{Chong Huang\\
University of California, Santa Barbara\\
{\tt\small chonghuang@umail.ucsb.edu}
\and
Yuanjie Dang, Peng Chen\\
Zhejiang University of Technology\\
{\tt\small dangyj@zjut.edu.cn, chenpeng@zjut.edu.cn}
\and
Xin Yang\\
Huazhong University of Science and Technology\\
{\tt\small xinyang2014@hust.edu.cn}
\and
Kwang-Ting (Tim) Cheng\\
Hong Kong University of Science and Technology\\
{\tt\small timcheng@ust.hk}
}

\maketitle

\begin{abstract}
Imitation learning has been applied to mimic the operation of a human cameraman in several autonomous cinematography systems ~\cite{chen2016learning, chen2015mimicking, chen2017should, huang2019cvprlearning}. To imitate different filming styles, existing methods train multiple models, where each model handles a particular style and requires a significant number of training samples. As a result, existing methods can hardly generalize to unseen styles. In this paper, we propose a framework, which can imitate a filming style by ``seeing" only a single demonstration video of the same style, i.e., one-shot imitation filming. This is done by two key enabling techniques: 1) feature extraction of the filming style from the demo video, and 2) filming style transfer from the demo video to the new situation. We implement the approach with deep neural network and deploy it to a 6 degrees of freedom (DOF) real drone cinematography system by first predicting the future camera motions, and then converting them to the drone's control commands via an odometer. Our experimental results on extensive datasets and showcases exhibit significant improvements in our approach over conventional baselines and our approach can successfully mimic the footage with a unseen style.
\end{abstract}

\section{Introduction}

Filming human motions with a camera drone is a very challenging task, because it requires the cameraman to manipulate the remote controller and meet the desired filming style simultaneously. Existing smart consumer drones make it more convenient by providing a few filming styles (e.g., orbiting mode in DJI spark) so that the drone can automatically fly along the predefined trajectory while capturing the video. However, this hard-copy auto filming only repeats the fixed patterns, which cannot satisfy the users' expectation. We expect the drone to be more intelligent to improvise cinematic videos like a human.

\begin{figure}[t]
\begin{center}
  \includegraphics[width=0.48\textwidth]{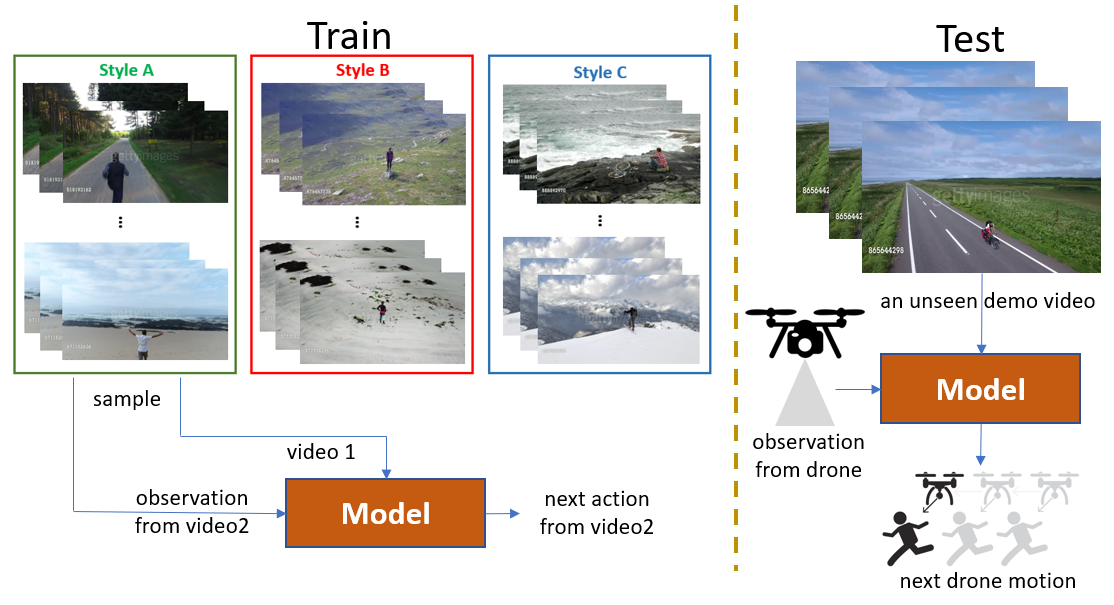}
  \caption{ We train a model which can imitate the filming style given a single video via learning to imitate a set of basic styles.}\label{intro}
\end{center}
\vspace{-0.5cm}
\end{figure}

Imitation-based auto filming can learn video shooting skills by imitating the camera operation from training data, such as ~\cite{chen2016learning, chen2015mimicking, chen2017should,huang2019cvprlearning}. However, existing imitation filming methods are style-specific. For example, the models in the work ~\cite{chen2016learning, chen2015mimicking, chen2017should} are trained to mimic a filming style where a moving subject is placed on the screen with the suitable image space in his/her moving direction. In the work ~\cite{huang2019cvprlearning}, a policy might have been trained through an imitation learning algorithm to orbit around the subject, and then another policy would be trained to chase the subject, etc. Each model has only a single-style control policy. In addition, these methods require significant amount of samples and training time for each style. If there are few video samples (even single sample) for one style, the learning model in the auto filming system will suffer from overfitting. This is far from what we desire. Ideally, we hope to demonstrate a certain filming style only one time to the robot, and have it instantly generalize to new situations of the same style without long training time. 

Several studies have been conducted in the literature for imitation learning from very few demonstration. Duan et al ~\cite{duan2017one} proposed a ``one-shot imitation learning" framework to enable the robotic to stack blocks as the desired height of the block towers given a single demonstration. Finn et al ~\cite{finn2017model} proposed a model-agnostic meta-learning (MAML) for better generalization performance on the same task. However, few of them are directly applicable to imitate the filming style in terms of the camera motion. 

On the one hand, existing one/few-shot imitation learning methods focus on learning the intent of one task from other similar tasks, while each task is defined based on the final state of the subject. For example, stacking blocks task is classified based on the final height of the block towers. In our filming task, filming style is related to the dynamic process of the entire video rather than the appearance of the last frame.  Therefore, these imitation learning methods can not guarantee that the model can learn to understand the filming style of the demo video only by minimizing action prediction error.

On the other hand, the conventional imitation learning methods requires the video and synchronized action as training data. However, our training data is collected from the website without the associated action variable. Although existing structure from motion techniques can estimate the camera trajectory, the ambiguous scale makes it infeasible to use the sequence of the camera pose to supervise learning the model and drive the camera motion.

In this work, we aim at an autonomous drone cinematography system (see Fig.~\ref{intro}), which can understand and imitate the filming style from only one demonstration via learning to imitate a set of basic styles. To this end, we define five basic styles (i.e., fly-through, fly-by, follow, orbiting, super-dolly), the mixture of which can cover most filming styles for videos including a single moving subject. We propose a one-shot imitation filming framework including a style feature extraction module and an action imitation module. We learn the filming style feature by minimizing the mis-classification error (of five predefined basic filming styles) of a classifier on the top of the feature extractor module. Because the filming style is jointly determined by the dynamic change of foreground and background, we design an attention-based network which takes the foreground feature (subject's on-screen position, size and orientation) and background feature (motion field) to capture the temporal information of the video content. The action imitation module predicts the camera motion in the following moments given three inputs: the style feature, current observation and camera motion. To reduce the difficulty in acquiring the ground-truth of 6DOF camera motion parameters from training videos, we utilize the combination of angular speed, linear speed direction and on-screen subject's size to represent control variable, which can be further converted as the actual motion during the online filming. In the test phase, if a demo video is a mixture of several basic style videos, we design a method to decompose the video into a sequence of single-style clips and imitate each clip orderly. Since there is no such data for the predefined basic styles, we collect a new dataset which contains 146 video clips including 5 styles. We analyze the impact of different inputs and compare the proposed method with several baselines. Experimental results demonstrate the superiority of our method to conventional baselines. We also deploy our model to a real drone platform and the real demo shows that our drone cinematography system can successfully achieve one-shot imitation filming. 

In summary, our contributions are four-fold:

\begin{itemize}
\item An imitation filming framework which could imitate the filming style from only a single demo video, and generalize it in a broader set of situations, greatly reducing the requirement for significant amount of samples and training time for each style.
\item An efficient method to represent the filming style for videos in terms of the temporal interaction between the subject and background while existing style representation methods do not consider the temporal ordering of frame sequences.
\item A new dataset consists of 146 video clips including 5 basic filming styles for learning one-shot imitation filming. The dataset will be released to benefit the community of learning-based filming.
\item Comprehensive experiments and ablation studies to demonstrate the superiority of the proposed method over the state-of-the-arts.  
\end{itemize}

We discuss related work in Sec. II, and introduce one-shot imitation filming in Sec. III. We detail the implementation of the algorithm in Sec. IV. We present the experimental results to evaluate our system in Sec. V, . Finally, we give the conclusion in Sec. VI.

\section{Related Work}

\textbf{Autonomous Aerial Filming}: Recent research works ~\cite{nagelireal, nageli2017real, galvane2016automated, galvane2018directing, joubert2016towards, kang2018flycam, huang2018through, huang2018act} enable flexible human-drone interactions in an aerial filming task. For instance, in ~\cite{nagelireal, nageli2017real, galvane2016automated, galvane2018directing, joubert2016towards, huang2018through, kang2018flycam}, the users are allowed to specify the subject size, viewing angle and position on the screen to generate quadrotor's motion automatically. However, these techniques essentially move a camera to a target pose based on the demand specified by users;  therefore, the aesthetic quality of the video highly relies on the user's input. Huang et al. ~\cite{huang2018act} designed an automatic drone filming system that estimate the next optimal camera pose which maximizes the visibility of the subject in an action scene. But only considering the subject's visibility is still too simplified to ensure a high-aesthetic quality of the captured video in various complex real-world scenarios. 

\textbf{Imitation Filming}: Imitation filming is essentially a data-driven autonomous camera planning solution. In ~\cite{chen2016learning} and ~\cite{chen2015mimicking}, the authors utilized video clips of basketball games to teach a camera to track a moving subject. In ~\cite{hu2017deep}, the authors learned a model based on images labeled with the object's positions for tracking the most salient object in a $360^{\circ}$ panoramic video.  Huang et al.~\cite{huang2019icralearning, huang2019cvprlearning} introduced machine learning techniques to teach a drone to imitate several predefined styles from aerial videos. However, these models are style-specific and each of them can only imitate a single style learned from the training data. 

\textbf{Filming Style Characterization}
Filming styles characterization have been well studied in multimedia community. Rath et al. ~\cite{rath1999iterative} proposed a four-parameter linear global motion model to describe the camera motion, i.e. pan and zoom. Bhattacharya et al.~\cite{bhattacharya2014classification} presented a discriminative representation of video shot which can effectively distinguish among eight cinematographic shot classes: aerial, bird-eye, crane, dolly, establishing, pan, tilt and zoom. Li et al.~\cite{li2017videography} constructed a videography dictionary to represent the foreground and background motion of each video clip. However, their video representation based on bag-of-visual-words does not consider the temporal ordering of video clips, so it cannot distinguish the long sequence (concatenation) of multiple different foreground/background motions.

\section{One-Shot Imitation Filming}

\subsection{Problem Formulation}
In this work, we focus on filming a single person in the scene. In particular,  the filming style in this work refers to the relative motion between the camera and the subject, therefore we can represent the filming style as a trajectory, which is parameterized to the time variable $t$ in each dimension of 6 DOF as follows: 

\begin{equation} 
\begin{aligned}
\label{action}
style & \sim \{a_t\}_{0:T} \\
a_{t} =  \{x^d_t-x^s_t, y^d_t-y^s_t, & z^d_t-z^s_t, roll^d_t, yaw^d_t, pitch^d_t\}
\end{aligned}
\end{equation} 

where $(.)^d$ and $(.)^s$ are the pose of the camera and subject in the world coordinates respectively, and $T$ is the duration of a video clip.

We aim to imitate a filming style by predicting the camera motion (i.e. $a_{t+1}$) of the next time ${t+1}$ that conforms to the style based on the observation $o_t$, the style information of the demo video $g_{\phi}(d)$ and the camera motion $a_t$ at time $t$,

\begin{equation} 
\begin{aligned}
\label{model}
a_{t} =  f_\theta(a_{t+1}|o_t,a_t,g_\phi(d))
\end{aligned}
\end{equation} 

Our objective is to learn two models $g_\phi()$ and $f_\theta()$, which extract representative style features of a demo video and accurate predict the next camera motion respectively.

Considering that in practice filming styles are usually a combination of several basic styles, we train a model to imitate a set of basic filming styles and for a demonstration video including arbitrary filming styles we imitate it based on combination of the learned basic styles. According to ~\cite{smith2016photographer}, five widely-used basic filming styles in single-subject drone cinematography include (see Fig.~\ref{styles}):

\begin{itemize}
\item Fly-through: pass through a subject without rotation.
\item Fly-by:  fly past the target in a straight line while rotating the camera to keep the subject framed in the shot.
\item Follow: follow the subject with the fixed distance.
\item Orbiting: rotate around the subject of interest while pointing to the subject.
\item Super-dolly: fly backwards, leading the subject. 
\end{itemize}

\begin{figure}[t]
\begin{center}
  \includegraphics[width=0.48\textwidth]{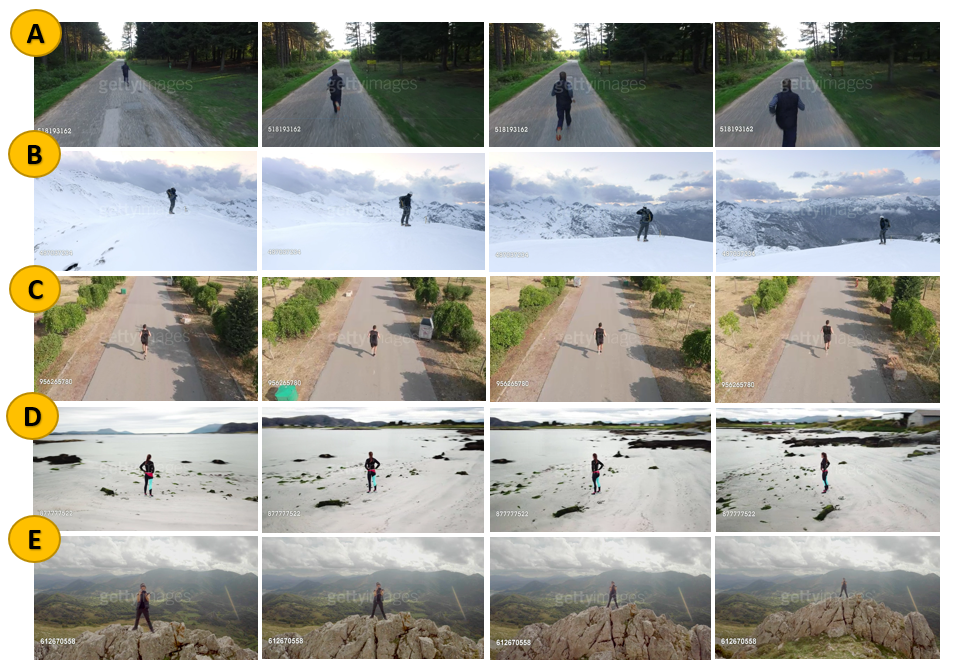}
  \caption{Exemplar videos with the style (A) fly-through (B) fly-by (C) follow (D) orbiting and (E) super-dolly. }\label{styles}
\end{center}
\vspace{-0.5cm}
\end{figure}

\subsection{Approach}
Although our imitation policy is similar to the conventional one-shot imitation learning, we do not follow their end-to-end training strategy to learn the entire model by minimizing the action prediction error. This is because different styles may correspond to similar sets of camera motion with different temporal orders. The action prediction error cannot be used to guide the model to learn style feature representation. Therefore, we train the two modules independently.

Because the mixture of five basic filming styles can cover most filming styles for videos, we train a style feature extraction module by minimizing the mis-classification error of five basic styles. Specifically, the style is related to the temporal interaction between the background motion and the foreground (i.e., subject) motion. The background/foreground embeddings and multi-feature fusion needs careful engineering.

To enable the agent to imitate the style from one video, we can allow it to incorporate prior experience, rather than learn each style completely from scratch. By incorporating prior experience, the agent should also be able to quickly learn to generalize the given style under different circumstances. To the end, we train the action imitation module based on the meta-learning method in ~\cite{duan2017one}. 

The above training process is based on the assumption that the input demo video has only a single basic style. In the test phase, the input demo video could be a long sequence (concatenation) of multiple basic styles, we need to analyze the motion coherency of the video and decompose the video into a sequence of single-style segments. Then we perform imitation filming for each segment orderly.

\begin{figure*}[t]
\begin{center}
  \includegraphics[width=0.9\textwidth]{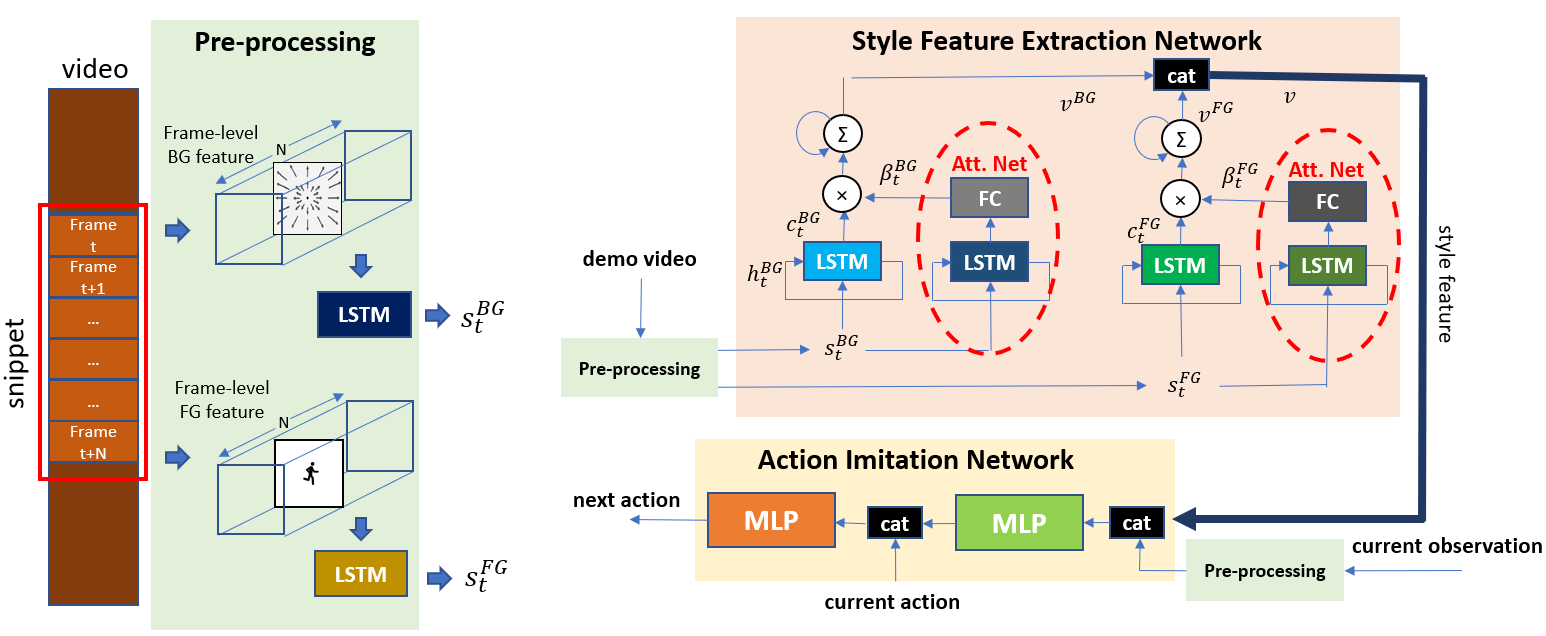}
  \caption{The proposed one-shot imitation filming framework. Pre-processing: we represent a video as a sequence of snippet embedding of foreground (FG) and background (BG). The style feature extraction network maps a variable-length demo video into a fixed-length style feature. The action imitation network predicts the next camera motion from the current observation, current action and style feature.}\label{architecture}
\end{center}
\vspace{-0.5cm}
\end{figure*}

\section{Implementation}
In this section, we introduce the implementation of the one-shot imitation filming framework. We choose the deep neural network to learn the mapping from the demo video and current state to the next action. Our proposed architecture mainly includes the style feature extraction network and action imitation network. To facilitate the network to extract the style-related information, we have an additional module to pre-process the video. An illustration of the architecture is shown in Fig. ~\ref{architecture}. We first introduce each module, then describe the camera estimation from the network outputs, and finally discuss how to handle the input video with multiple styles. 

\subsection{Pre-processing}

The input image sequence is downsampled as 4fps to reduce the duplicate information. We represent the foreground and background as follows:

\textit{Background}: We adopt the grid-based motion field ~\cite{li2017videography} to describe the motion between adjacent frames. In details, we utilize densely computed Kanade-Lucas-Tomasi (KLT) tracks ~\cite{tomasi1991tracking} over the entire sequence. The image plane is splitted as a K $\times$ L regular grids. We calculate the per-grid velocity $V_{k{\times}l}$ as the average from multiple KLT tracks intersecting that grid. As suggested in ~\cite{li2017videography}, the block size is set to 8$\times$8. 

\textit{Foreground}: We focus on the subject's appearance in terms of the on-screen position, size and upper-body orientation. The position and size can be described as a four-dimension bounding box, and the upper-body orientation is represented as a one-dimension Euler angles.

Inspired by the fact that snippet embedding in the action recognition ~\cite{wang2016temporal} can enhance the temporal representation, we adopt a overlapping sliding window to segment the video as multiple snippets and encode the background/foreground feature within each snippet. More specifically, Within a sliding window including $N$ background/foreground feature, we utilize an auto-encoder ~\cite{srivastava2015unsupervised} based on Long Short Term Memory (LSTM) networks ~\cite{xingjian2015convolutional} to learn embedding of the feature sequence. We train two encoders for background and foreground feature respectively. As the sliding window scans the entire video, a video can be represented as a sequence of combination of the background embedding and foreground embedding. In this work, we set $N$ as 8 for its maximum performance.

\subsection{Style Feature Extraction Network}
The style feature extraction network takes a sequence of background embedding and foreground embedding as input, and produces a fixed-length embedding of the video style to be used by the action imitation network.

Since our neural network needs to handle demo videos with variable lengths, LSTM network is a natural operation due to its ability to map the variable-dimensional inputs to fixed-dimensional outputs. Considering that the background and foreground contribute to the style classification differently, the backbone is consisted of two parallel main networks, which process the background and foreground embedding, respectively. 

In addition, the amount of valuable information provided by different frames is in general not equal. Only some of the frames (key frames) contain the most distinctive information about the style. For example, for the style ``fly-through", the snippet captured when the camera is passing by the subject should have higher importance than the snippet when the camera is moving closer to the subject. Based on such insight, we design a temporal attention network to automatically pay different levels of attention to different snippets. Similarly, we apply two parallel temporal attention networks to process the foreground and background independently. 

Let $c_t^{FG/BG}$ and $\beta_t^{FG/BG}$ denote the outputs of the foreground/background main networks and the temporal attention value of the foreground/background attention network at each time step $t$, the style feature $v$ is the concatenation of the weighted summation of the outputs at all time steps $T$:

\begin{equation} 
\begin{aligned}
\label{style_feature}
v = [\sum_{t=0}^{T} \beta_t^{FG} \cdot  c_t^{FG}: \sum_{t=0}^{T} \beta_t^{BG} \cdot c_t^{BG}] 
\end{aligned}
\end{equation}

We train the network by feeding the style feature into a 5-dim fully-connected layer $p$ followed with a softmax layer to convert to probability of five basic styles. The loss function is written as Eq.~\ref{loss_vu},  including 1) the cross-entropy loss of style mis-classification 2) the L2-norm loss of the foreground temporal attention and 3) the L2-norm loss of the background temporal attention. 

\begin{equation} 
\begin{aligned}
\label{loss_vu}
\min_{\phi} & \quad \sum_{c=1}^{C}  y_c log p_{c}(g_{\phi}(s_{1:T}^{FG},s_{1:T}^{BG})) \\
&+ \frac{\lambda_{1}}{T} \sum_{t=1}^{T}\| \beta_t^{FG}\|_{2} +  \frac{\lambda_{2}}{T} \sum_{t=1}^{T} \|\beta_t^{BG} \|_{2}  \\
\end{aligned}
\end{equation} 

where $C$ is the number of the basic styles (i.e. $C$=5) and $T$ is the number of the snippets of each video. In this work, we set both of $\lambda_{1}$ and $\lambda_{2}$ as 0.01.

\subsection{Action Imitation Network}
We aim at a learner which can predict the next camera motion under different contexts (the actual observation) so that the captured video matched the style of the demo video. The action imitation network takes the style feature from the demo video, the current observation and camera motion as inputs and predicts the next camera motion. We embed the latest $N$ frames from the camera as the foreground and background embeddings. 

Meanwhile, because it is impossible to derive absolute camera motions directly from training videos which do not contain inertial sensor data collected from the internet, we apply a simple and efficient method to represent the necessary control variable for our applications. First, we utilize the structure from motion techniques to estimate the camera trajectory (without the scale) from the video. Because each camera pose in the estimated trajectory has one timestamp, we can obtain the angular speed and linear velocity direction in the world coordinates. Second, because the style is essentially determined by the relative position between the subject and the camera, we can use the subject's height to represent the scale of the relative camera motion (if we assume that subject's height would not be changed frequently). Therefore, we use the angular speed, linear velocity direction and the subject's on-screen height to approximate the camera motion relative to the subject as the output label of the network. It is noted that the subject's on-screen height is normalized between 0 and 1 by dividing the pixel-wise the subject's height by the height of the image screen.

The network is constructed by two multilayer perception (MLP) in concatenation. The first MLP produces a context embedding given the style feature, the foreground and background embeddings, and the second MLP predicts the next camera motion from the context embedding and the current camera motion. 

Ideally, a context embedding should include the style-related information from the observation and ignore the trivial factors (e.g. background layout). Although different videos with the same style are different in terms of the camera speed, background layout and the subject's motion, they share the same trend of the temporal appearance change. Therefore, there exists the temporal matches (the same development progress) between two videos with the same style. Based on this insight, we propose a multi-task learning strategy to enhance the representation of the context embedding: First we sample two videos (style video and content video) with the same style. We sample a snippet (content snippet) from the content video and find the matching snippet (style snippet) from the style video. The context embedding, which is calculated from the style feature of style video and the content snippet, should  predict the next action for the content snippet and style snippet respectively (conditioned on the current action in the content and style snippets). We perform the same training procedure for 5 basic styles, and the loss function can be written as follow:

\begin{equation} 
\begin{aligned}
\label{loss_ai}
\min_{\theta} \quad  \| f_{\theta}(a_t^c, v, o_t^c) - a_{t+1}^{c*} \| + \lambda* \| f_{\theta}(a_{t'}^s, v, o_t^c) - a_{t'+1}^{s*} \|
\end{aligned}
\end{equation} 

where the $(.)^s$ and $(.)^c$ refer to the variable of style snippet and content snippet. $(.)^*$ indicates the ground-truth camera motion. $t$ and $t'$ are two matching timesteps from two videos. The hyper-parameter $\lambda$ is set as 0.7.

\begin{figure}[t]
\begin{center}
  \includegraphics[width=0.48\textwidth]{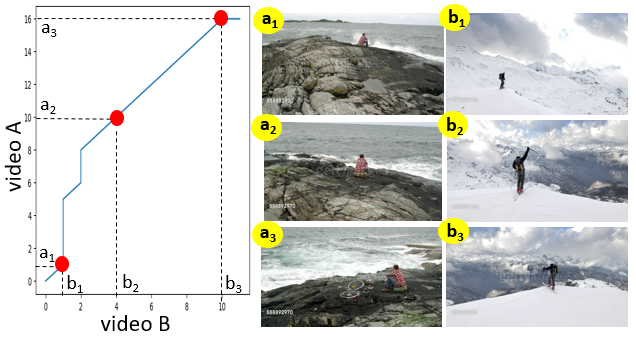}
  \caption{ Left: the warping path generated from two videos with the same style. Right: the matching snippets in each row share the similar relative subject's position.}\label{dtw}
\end{center}
\vspace{-0.5cm}
\end{figure}

\textbf{How to find the matching snippets?} If we consider the video as a sequence with variable length, there are multiple ways to find the optimal matching pairs between two time series. In this work, we utilize the dynamic time warping (DTW) ~\cite{berndt1994using} to detect the matching snippet of two videos, while the video is represented as a sequence of concatenation of background and foreground embeddings. Fig.~\ref{dtw} shows that the warping path in which the two sequences with style ``fly-by" are aligned in time. We can see that each red point on the warping path corresponds to two matching snippets $a_i$ and $b_i$, which share the similar relative subject's position.

\subsection{Camera Motion Estimation}

\begin{figure}[t]
\begin{center}
  \includegraphics[width=0.48\textwidth]{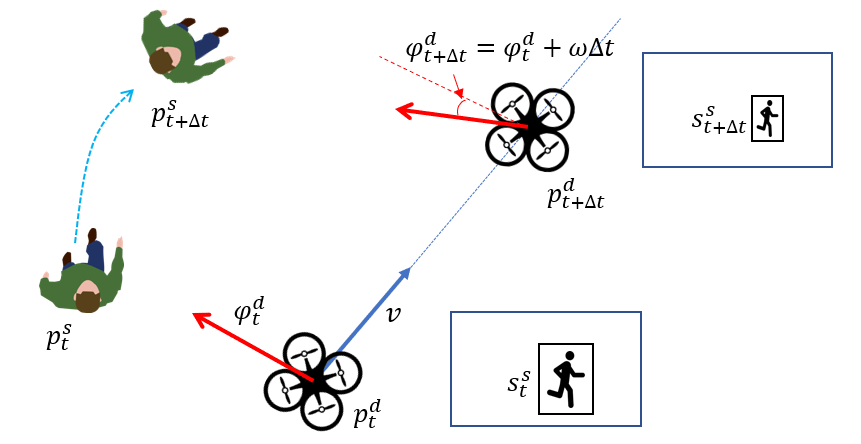}
  \caption{ The camera dynamically estimates the next camera pose based on its actual observation and the network prediction.}\label{control}
\end{center}
\vspace{-0.5cm}
\end{figure}

This section describes how to produce the camera motion from the outputs (angular speed $\omega$, linear velocity direction $v$ and subject's scale $s$) of the prediction network during online filming. In details, we have the drone's position $p^d_{t}$ and orientation ${\varphi}^d_{t}$ at the timestep $t$. If the subject's height is known, we can utilize Lim et al's method ~\cite{lim2015monocular} to localize the subject's position $p^s_{t}$ based on its bounding box  $s^s_{t}$. We assume that the subject's movement is smooth and we can use Kalman Filter to predict the subject's location $p^s_{t+{\vartriangle}t}$ in the next timestep. As Fig.~\ref{control} shows, the drone's orientation ${\varphi}^d_{t+{\vartriangle}t}$ in the next timestep can be obtained by adding ${\varphi}^d_{t}$ with ${\omega}{\vartriangle}t$. The drone's position ${p}^d_{t+{\vartriangle}t}$ in the next step is calculated by searching a position on the ray with the direction $v$ to minimize the error between the observed bounding box ${s}^s_{t+{\vartriangle}t}$ on this position and the predicted subject's scale $s$. This estimated waypoint (${\varphi}^d_{t+{\vartriangle}t}$ and ${p}^d_{t+{\vartriangle}t}$) will be sent to the actuator to control the drone.

\subsection{Video Segmentation}
\begin{figure}[h]
\begin{center}
  \includegraphics[width=0.45\textwidth]{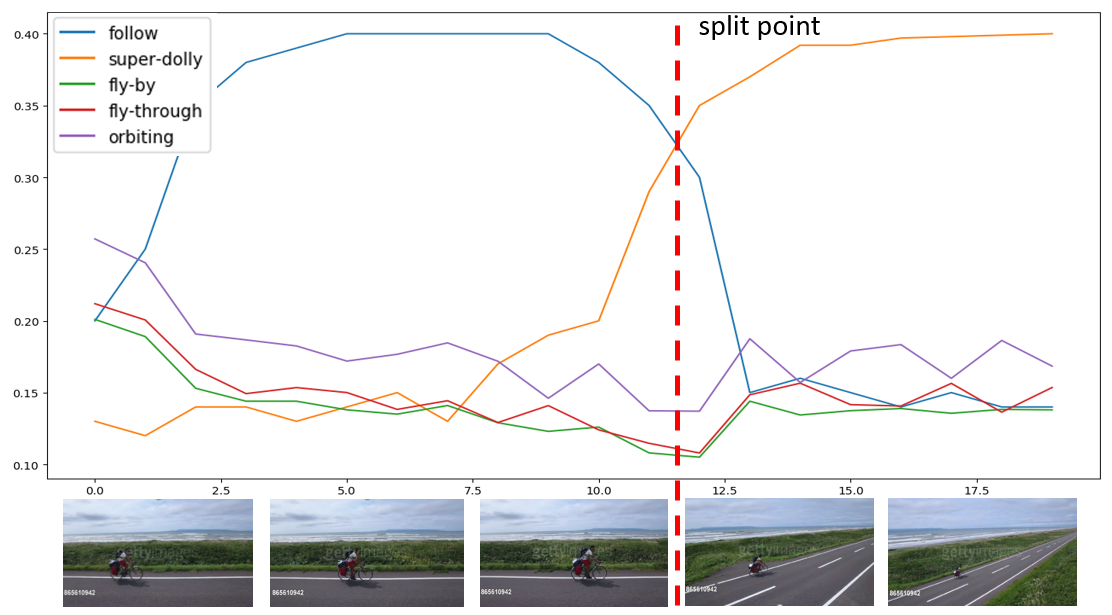}
  \caption{Curve of the prediction probability of a video along the time (increasing frames). The red dash line divides the video into two parts with styles ``follow" and ``super-dolly".}\label{split}
\end{center}
\vspace{-0.5cm}
\end{figure}

Because the network is trained based on the video clip with the single style, we need to handle the situation where the input demo video includes multiple basic styles in the test phase. We assume that the video can be considered as the concatenation of multiple variable-length clips, each of which only has a single basic style, so we can segment the video into a sequence of single-style video clips and perform imitation of each individual clip orderly. To define the segmentation rule, we calculate the style probability from the 5-class classifier of a video with the increasing frames. As Fig.~\ref{split} shows, the demo video can be visually divided as two parts with the styles: follow ($\le$11s) and super-dolly($\ge$11s). We observe that the continuous frames with the same style (follow) would increase the style probability until saturation. Once the video is transferring to a different style (super-dolly), the probability of the``follow" style  begins to decrease while the probability of the ``super-dolly" style  increases and exceeds the one of ``follow" at the time 11s. Based on this observation, our segmentation method is composed of two steps that are iteratively repeated: 1) We feed a video into the classifier frame by frame and calculate the curve of prediction probability of 5 basic styles. 2) We cut the video until that the probability of the original major style decreases by a threshold (we set it as 0.6), and then reset the network state and repeat the first step.

\section{Experiments}
In this section, we first introduce the dataset collection, and then describe the experimental setup and the measurement metrics, followed by experimental results.

\subsection{Dataset} 

We collect the video clips from the website  \textit{www.gettyimages.com}, which offers professional photography and videography. Specifically, we used three keywords ``\textit{aerial view, one person only, outdoor}'' to initialize our search. We excluded the searched video results which contain extremely poor lighting conditions, subjects taking up too small regions and/or being occluded for too long time during the video. 

To select the video with the predefined basic styles , we recruit 3 human annotators and asked them to manually label the videos based on the definition of each style. Each video was labeled by 1 annotator and verified and corrected by the other 2 annotators. We will drop the video if it does not belong to any of the basic styles. Eventually, we obtain 146 videos, each of which is around 5-50 seconds long, yielding videos of totally 3218 seconds. Tab.~\ref{styles} shows the statistics of the style annotations in our data.

We resized each video frame to 640x480 and down-sampled the video to frame rate of 4fps to adapt to the actual computation speed. In addition, we provide the ground-truth of camera trajectory (rotation and translation) and subject on-screen information (position, size and orientation). More specifically, we apply the state-of-the-art structure from motion tool OpenSFM ~\cite{opensfm} to extract the ground-truth of camera trajectory. The ground-truth of subject on-screen bounding box is detected based on YOLOv3 ~\cite{redmon2018yolov3} and the orientation is estimated based on the result of ~\cite{doe2019}, while we manually correct the misidentified skeleton joints to replace the original result.

\begin{table}[h]
\footnotesize
\caption{Statistics of the style annotations in our data}
\label{styles}
\begin{center}
\begin{tabular}{|c|c|c|c|c|c|}
\hline
Style & fly-by & fly-through & follow & orbiting & super-dolly \\
\hline
Videos & 21 & 42 & 30 & 28 & 25\\
\hline
\begin{tabular}{@{}c@{}}Duration \\ (Second)\end{tabular} & 452  & 976   & 670  & 587  & 533 \\
\hline
\end{tabular}
\end{center}
\vspace{-0.5cm}
\end{table}

\subsection{Experimental Setup}

We split our dataset into 97 training videos and 49 test videos. The number of videos from the five styles (i.e. fly-by, fly-through, follow, orbiting, super-dolly) are 14, 28, 20, 18, 17 for the training set and 7,14,10,10,8 for the testing set. For each training and testing video, we applied an overlapping sliding window with a length of 8 to generate a set of snippets for background/foreground embedding. The stride of the overlapping sliding window is 4. Accordingly, we generate a total of 1960 training clips and 1002 testing clips. We further augmented the training data by flipping each video clip along the horizontal axis, yielding 3920 training clips. We train our network on the Nvidia Tesla K50c and utilize Adamax ~\cite{kingma2014adam} to perform the optimization, with a learning rate of 0.001.

We evaluate the performance of our method using three types of metrics: 

1) The confusion matrix of the style classification which aims at examining the representation ability of the style feature in terms of style classification. Meanwhile, we also use it to evaluate imitation performance against the demo video.

2) The mean square error (MSE) which measures the differences between the predicted camera motion and the ground-truth in terms of angular speed ($rad/s$), linear velocity direction and ($rad$) the subject's normalized scale.

3) A subjective quality score obtained from a user study. We recruited 10 volunteers and each volunteer was asked to score the similarity between the recorded video with training videos (from 1: worst to 5: best). We calculate the average score of the 10 volunteers for each testing video and then average the score on all testing videos.

\subsection{Style Representation}

In this subsection, we design the experiments based on our dataset to carefully analyze two factors on the style classification results: 1) input feature selection and 2) attention mechanism. To the end, we design the following baselines: FG-only, BG-only, FG+BG, FG+BG+Att. Details of the four baselines are listed in Tab.~\ref{baselines}.

\begin{table}[h]
\caption{Design of the four network baselines}
\label{baselines}
\footnotesize
\begin{center}
\begin{tabular}{|c|c|c|c|c|}
\hline
 & FG-only & BG-only & FG+BG & FG+BG+Att \\
\hline
foreground & \checkmark  & & \checkmark  & \checkmark  \\
\hline
background & &  \checkmark  & \checkmark  & \checkmark  \\
\hline 
attention & & &  & \checkmark  \\
\hline
\end{tabular}
\end{center}
\vspace{-0.5cm}
\end{table}

\begin{figure}[h]
\begin{center}
  \includegraphics[width=0.5\textwidth]{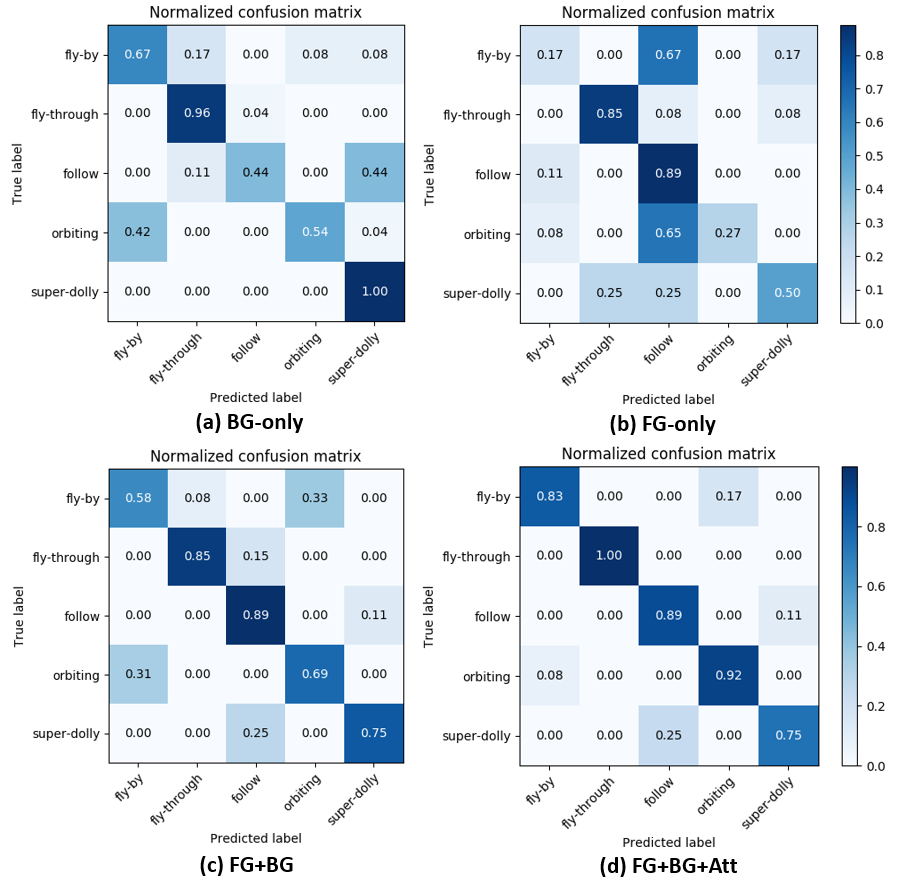}
  \caption{The confusion matrix of four baselines.}\label{confusion_matrix}
\end{center}
\vspace{-0.5cm}
\end{figure}

\begin{figure}[h]
\begin{center}
  \includegraphics[width=0.5\textwidth]{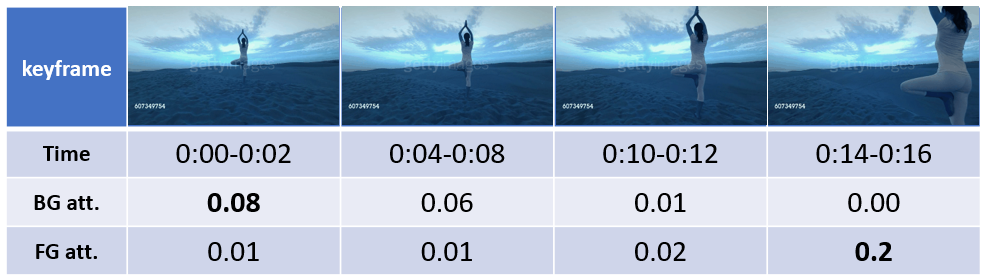}
  \caption{The foreground/background attention weights (FG/BG att.) of a video with the style ``fly-through".}\label{attentions}
\end{center}
\vspace{-0.5cm}
\end{figure}

Because FG-only and BG-only baselines only use one branch network, we double the number of hidden neurons such that they have the same number of parameters as other two baselines. Fig.~\ref{confusion_matrix} shows that the foreground and background have complementary relationship for filming styles classification. For example, the orbiting style depends more on the background than the foreground, while it is reverse for the follow style. In addition, the baseline FG+BG performs better than the baselines with a single input (FG-only and BG-only), indicating that the combination of both feature can further improve the classification performance. The baseline FG+BG+Att achieves the best performance among all the baselines, which proves that the attention layer is beneficial to process long sequence.

To further investigate where the attention layer focus on, we visualize the attention weights of the background and foreground. Fig.~\ref{attentions} shows the distribution of the attention weights of a video with the style ``fly-through". We can see that when the subject's on-screen size becomes large, the network pays more attention on the foreground. Specifically, the network assigns the highest weight to the clip during time interval 0:14-0:16 because this 2-second clip is much more distinctive than other clips in terms of style recognition. 

\subsection{Action Prediction}
In this subsection, we compare the imitation performance of the models trained by the proposed loss function (Eq.~\ref{loss_ai}) and the loss in ~\cite{duan2017one}. Tab.~\ref{dual_loss} shows the prediction accuracy of the angular speed ($\omega$), linear direction vector ($v$) and the scale ($s$) in terms of different filming styles. We can see that the our proposed method can keep consistent improvement over the model trained from the method ~\cite{duan2017one} in different styles.

\begin{table}[h]
\caption{Comparison of action prediction with different loss}
\label{dual_loss}
\small
\begin{center}
\begin{tabular}{|c|c|c|c|c|c|c|}
\hline
 & \multicolumn{3}{c|}{proposed method} & \multicolumn{3}{c|}{Duan et al. ~\cite{duan2017one}} \\
\hline
style & $\omega$ & $v$ & $s$ & $\omega$ & $v$ & $s$\\
\hline
fly-by & 0.04 & 0.43 & 0.04 & 0.07 & 0.86 & 0.04\\
\hline
fly-through & 0.01 & 0.01 & 0.02 & 0.01 & 0.01 & 0.03 \\
\hline
orbiting & 0.04 & 0.03 & 0.02 & 0.06 & 0.04 & 0.02\\
\hline
follow & 0.01 & 0.00 & 0.02 & 0.01 & 0.01 & 0.02\\
\hline
super-dolly & 0.01 & 0.02 & 0.02 & 0.01 & 0.02 & 0.03\\
\hline
\end{tabular}
\end{center}
\vspace{-0.5cm}
\end{table}

\subsection{Application to Drone Cinematography System}

In this subsection, we deploy our one-shot imitation filming method to a real drone platform for the autonomous cinematography task. Specifically, we build our drone cinematography system on the DJI Matrix 100 with two onboard embedded systems (Nvidia Jetson TX2 and DJI Manifold). We feed a demo video and captured a new video within the same duration.

First we evaluate imitation filming of the demo video with a single basic styles. We capture 5 videos for each style and feed the style feature of each video into classifier network.  We utilize the style classification accuracy and the user study (as described in Sec.4.1) to evaluate the imitation performance. Tab.~\ref{single_style} shows that our network can accurately imitate the demo video with the single basic filming style.

\begin{table}[h]
\caption{Comparison of imitation performance of basic styles}
\label{single_style}
\tabcolsep=0.13cm
\footnotesize
\begin{center}
\begin{tabular}{|c|c|c|c|c|c|}
\hline
style & fly-by & fly-through & follow & orbiting & super-dolly\\
\hline
accuracy & 0.8  & 1.0 & 1.0 & 1.0 & 1.0 \\
\hline
user-study & 4.0$\pm$0.2  & 4.5$\pm$0.5 & 4.7$\pm$0.3 & 4.6$\pm$0.4 & 4.8$\pm$0.2 \\
\hline
\end{tabular}
\end{center}
\vspace{-0.5cm}
\end{table}

Second, we evaluate our method in terms of imitating 5 videos with the mixed styles. Meanwhile, we also compare our system with the state-of-the-art autonomous drone cinematography system ~\cite{huang2019cvprlearning}, which learns each filming style completely from scratch. The user study in Tab.~\ref{mixed_style} shows that our system achieves better performance than the system ~\cite{huang2019cvprlearning}. This is because that the imitation filming framework in the system  ~\cite{huang2019cvprlearning} suffers from overfitting (the model is trained only from the given demo video). The attached videos will provide a more convincing comparison.

\begin{table}[h]
\caption{Comparison of imitation performance of the mixed styles}
\label{mixed_style}
\footnotesize
\begin{center}
\begin{tabular}{|c|c|c|}
\hline
style & proposed & Huang et al. ~\cite{huang2019cvprlearning} \\
\hline
user-study & 4.2$\pm$0.6  & 2.4$\pm$0.5 \\
\hline
\end{tabular}
\end{center}
\vspace{-0.5cm}
\end{table}

\section{Conclusions}
We propose a novel and efficient filming framework one-shot imitation filming, where the camera agent can imitate a filming style by ``seeing" only a single demonstration video of the same style. The proposed framework comprises two modules: 1) style feature extraction, and 2) action imitation. Compared with the state-of-the-art imitation filming techniques, our method does not require significant amount of samples and training time for each style. Our experimental results on the datasets and showcases exhibit significant improvements of our approach over existing methods and our approach can successfully mimic the footage with a unseen style.

{\small
\bibliographystyle{ieee_fullname}
\bibliography{egpaper_final}
}

\end{document}